\documentclass[letterpaper, 10 pt, conference]{ieeeconf}  

\IEEEoverridecommandlockouts                              

\usepackage{graphics} 
\usepackage{graphicx}
\usepackage{cuted}
\usepackage{capt-of}
\usepackage{epsfig} 
\usepackage{mathptmx} 
\usepackage{times} 
\usepackage{amsmath} 
\usepackage{amssymb}  
\usepackage{booktabs}
\usepackage{array}
\usepackage{graphicx}
\usepackage{tabularx}
\usepackage{siunitx}
\usepackage{cite}
\usepackage{todonotes}
\usepackage{hyperref}
\usepackage{stfloats}

\usepackage{fontawesome5}
\usepackage[most]{tcolorbox}
\newtcbox{\linkpill}{
  on line,
  box align=base,
  colback=gray!10,
  colframe=gray!10,
  arc=1.5mm,
  boxrule=0pt,
  left=2pt,
  right=2pt,
  top=1pt,
  bottom=1pt
}
\definecolor{slsorange}{HTML}{D97706}
\definecolor{slsnavy}{HTML}{174A7E}

\definecolor{mumpccyan}{HTML}{50D6FF}
\newtcbox{\titlemu}{
  on line,
  box align=base,
  colback=mumpccyan!6!white,
  colframe=mumpccyan,
  coltext=mumpccyan,
  arc=.21em,
  boxrule=.019em,
  boxsep=0pt,
  left=.18em,
  right=.18em,
  top=.12em,
  bottom=.12em
}

\AtBeginDocument{
    \addtolength{\abovedisplayskip}{-0.4em}
    \addtolength{\belowdisplayskip}{-0.4em}
    \addtolength{\abovedisplayshortskip}{-0.4em}
    \addtolength{\belowdisplayshortskip}{-0.4em}
}

\makeatletter
\def\subsubsection{\@startsection{subsubsection}{3}{\z@}%
  {0.5\baselineskip plus 0.2\baselineskip minus 0.1\baselineskip}%
  {0em}%
  {\normalfont\normalsize\itshape}}
\makeatother

\title{\LARGE \bf
$\mu$MPC: Microsecond-Scale Nonlinear and World Model Control \\ from Edge to GPU via Koopman-Linearized Precompilation
}

\author{Devesh Nath\and Glen Chou
\thanks{The authors are with the Georgia Institute of Technology, Atlanta, GA 30308, USA. \texttt{\{\href{mailto:dnath7@gatech.edu}{dnath7},\href{mailto:chou@gatech.edu}{chou}\}@gatech.edu}}%
}

\begin{document}

\maketitle
\thispagestyle{empty}
\pagestyle{empty}

\begin{figure*}[!b]
    \centering\vspace{-15pt}
    \includegraphics[width=\textwidth]{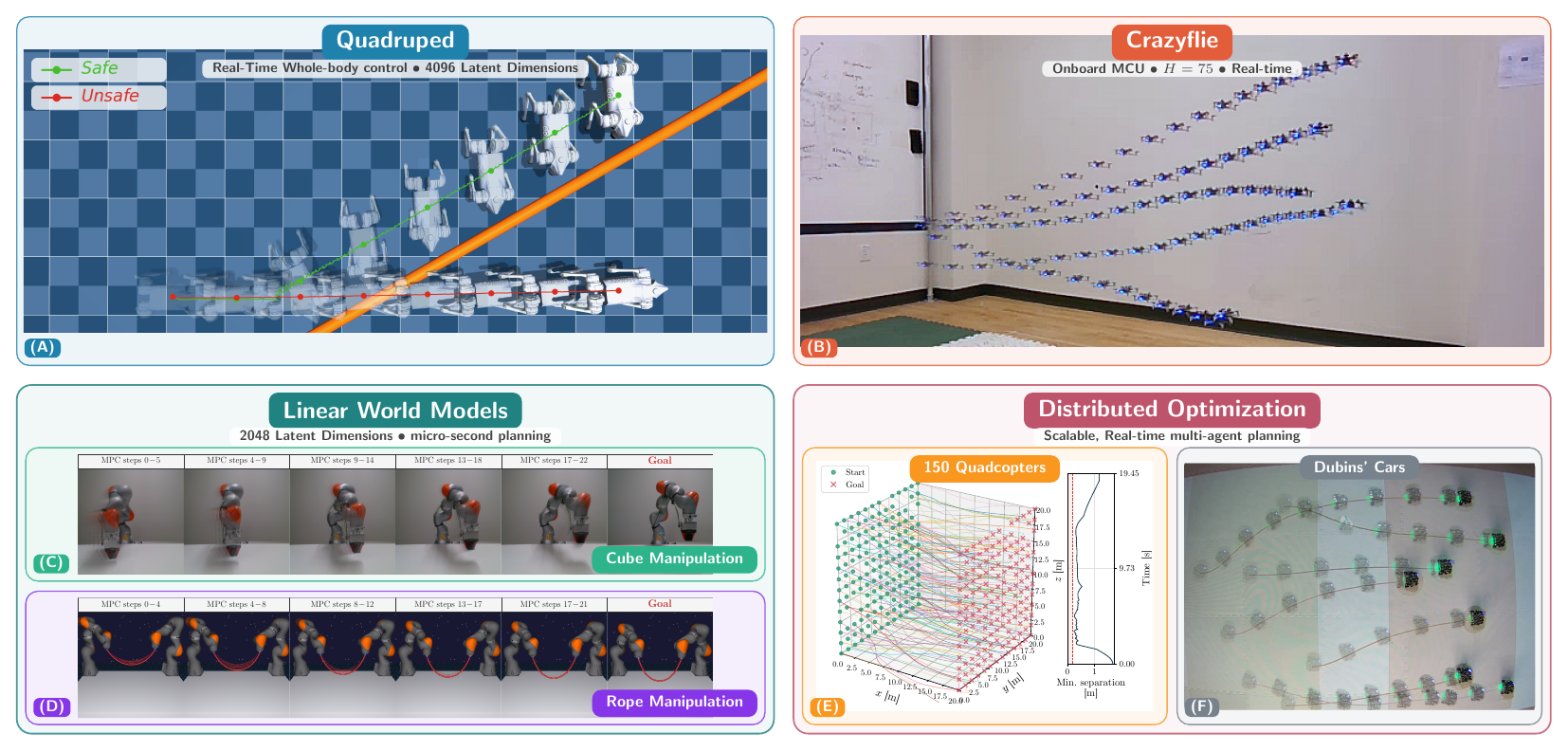}
    \vspace{-1.5em}
    \captionof{figure}{
    \textbf{$\mu$MPC} enables real-time control from microcontrollers to massive learned models and multi-agent
    systems: \textbf{(A)} quadruped safety filtering with a 4096-dimensional
    (${\sim}16$M-parameter) Koopman model; \textbf{(B)} onboard $H=75$ Crazyflie MPC;
    \textbf{(C--D)} hardware cube and rope manipulation; \textbf{(E)} collision-constrained planning for
    150 quadrotors; and \textbf{(F)} seven-robot multi-agent planning on the Robotarium.
    }
    \label{fig:front_page} \vspace{-15pt}
\end{figure*}

\begin{abstract}
\looseness=-1
We present $\mu$MPC, a fast model predictive control (MPC) framework that reduces online computation using learned Koopman-linearized dynamics. By precomputing finite-horizon control-to-output maps, $\mu$MPC eliminates lifted-state trajectories, yielding an action-only Koopman-Compiled QP (KC-QP) with $Hn_u$ decision variables, independent of lifted dimension. A fixed, precomputed primal operator enables efficient primal-dual iterations using only matrix-vector products and projections, accommodating changing goals and constraints without refactorization. This efficiency also enables control-aware dynamics learning through differentiable KC-QPs in the training loop and batched DAgger-style data aggregation. $\mu$MPC achieves solve times as low as $23\mu$s on a desktop CPU and 24 ms onboard a 168 MHz Crazyflie microcontroller. It scales to 4096D quadruped control in under 5 ms, visual planning with 2048D world models in 0.4 ms, and collision-aware planning for 200 quadrotors at over 30 Hz on the GPU. Across these benchmarks, $\mu$MPC achieves up to $108\times$ faster constrained MPC and $1907\times$ faster visuomotor planning than competing methods. 
\end{abstract}
\begin{center}
\vspace{-5pt}
{\linkpill{\href{https://www.deveshnath.com/projects/micrompc/}{\textcolor{slsnavy}{\faGlobe\ {\textbf{Project Website}}}}}
\ \ 
\linkpill{\href{https://github.com/trustworthyrobotics/microMPC}{\textcolor{slsnavy}{\faGithub\ {\textbf{Code}}}}}\ \ 
\linkpill{\href{https://youtu.be/KMsVFYA1S3A}{\textcolor{slsnavy}{\faYoutube\ {\textbf{Video}}}}}}
\vspace{-8pt}
\end{center}

\section{Introduction}

\looseness=-1
Optimal control of complex robotic systems requires expressive dynamics models and real-time solvers for deployment, which is particularly difficult for resource-constrained and high-dimensional systems. Embedded aerial robots face strict compute, memory, and power budgets \cite{nguyen2024tinympc, du2026gato}; whole-body legged control requires high-dimensional real-time optimization \cite{amatucci2026primaldual, fang2026gpusls}; visuomotor planning relies on learned world models with thousands of latent states \cite{zhou2025dinowm, nath2026pixels}; and multi-agent systems require many concurrent optimizations \cite{zhan2026robustly, foderaro2014distributed}. Controllers must also plan over long horizons, react rapidly to observations, and satisfy complex constraints. Nonlinear model predictive control (MPC) addresses these requirements, but its computational cost grows with problem dimension, horizon length, and constraint complexity. Despite advances in sparsity exploitation, factorization caching, and GPU parallelism \cite{fang2026gpusls, amatucci2026primaldual}, real-time deployment remains challenging. 

\looseness=-1
This computational challenge is particularly pronounced for learned dynamics and world models. While these models can capture complex dynamics from high-dimensional observations, planning often requires repeated computationally-expensive evaluations of candidate action sequences through the learned dynamics \cite{zhou2025dinowm, maes2026leworldmodel, gao2026fastlewm}. One approach to reducing this cost is via \textit{Koopman linearization}, which learns dynamics that are linear in a lifted space, yielding an affine dependence of finite-horizon predictions on the control sequence \cite{korda2018linear} that simplifies both prediction and trajectory optimization. However, nonlinear constraints can still require iterative linearization and expensive matrix refactorizations, limiting the computational benefits. This motivates an MPC framework that moves expensive computations offline while allowing goals and constraints to be updated efficiently online.

\looseness=-1
Another challenge is learning dynamics models that are suitable for control rather than prediction alone. Conventional Koopman dynamics learning emphasizes state-prediction accuracy \cite{mezic2026koopmansurvey}, but low prediction error may not translate to high control performance. Moreover, training on fixed demonstrations neglects controller-induced distribution shifts, potentially degrading prediction accuracy at inference time. Differentiable MPC offers a way to address these limitations \cite{amos2017optnet} by incorporating the learned Koopman dynamics into the MPC optimization during training, enabling direct supervision of control performance and generating training data arising from controller-induced state distributions. However, repeatedly solving and differentiating through MPC problems can be prohibitively expensive during training.


\looseness=-1
To address these challenges, we present $\mu$MPC, an MPC framework for nonlinear systems that exploits learned Koopman-linearized dynamics to formulate Koopman-Compiled Quadratic Programs (KC-QPs) that can be solved in microseconds. We develop a tailored primal-dual solver that precomputes a fixed primal operator offline, reducing online constrained optimization to efficient matrix-vector products and projections. This operator can be reused across changing constraints, goals, and MPC steps, substantially reducing online computation. The efficiency of KC-QPs also enables scalable control-aware learning: we solve batches of KC-QPs in parallel, evaluate the resulting control performance under the original nonlinear dynamics, and differentiate through the optimization to train Koopman representations directly for control performance rather than prediction accuracy alone. Furthermore, fast batched KC-QP solves enable DAgger-style on-policy data aggregation, improving coverage of controller-induced state distributions and mitigating inference-time distribution shift. Together, this scalable learning of high-dimensional Koopman representations tailored to closed-loop control enables microsecond-scale online MPC. 
Our contributions are:
\begin{itemize}

\looseness=-1
\item $\mu$MPC exploits the linear structure of learned Koopman dynamics to eliminate state variables and formulate MPC as an action-only QP (KC-QP). We precompute a fixed primal operator offline, enabling fast, ``precompiled" primal-dual iterations using only matrix-vector products and projections, even as goals and constraints change, without requiring online matrix refactorization.

\looseness=-1
\item We exploit the compact, parallelizable structure of KC-QPs to train the Koopman dynamics to produce MPC control sequences that match expert demonstration data, by differentiating through the optimization. Batched KC-QP solves also enable fast DAgger-style data aggregation, reducing controller-induced distribution shift.

\looseness=-1
\item We demonstrate $\mu$MPC across embedded, high-dimensional, multi-agent, and visual control. Our solver achieves MPC runtimes as low as 23 $\mu$s on Dubins car and quadrotor systems, solves $H=75$ MPC onboard a resource-limited Crazyflie microcontroller in 24 ms, and performs 4096-dimensional quadruped control in under 5 ms. We further demonstrate real-time collision-aware multi-agent planning for 200 quadrotors, with hardware validation on a system of seven differential-drive robots. For visual control, we train 2048D linear world models enabling sub-millisecond planning, outperforming evaluated baselines on navigation and bimanual rope manipulation, with hardware cube-manipulation validation. 

\end{itemize}

\begin{figure*}[t]
    \centering
    \includegraphics[width=\textwidth]{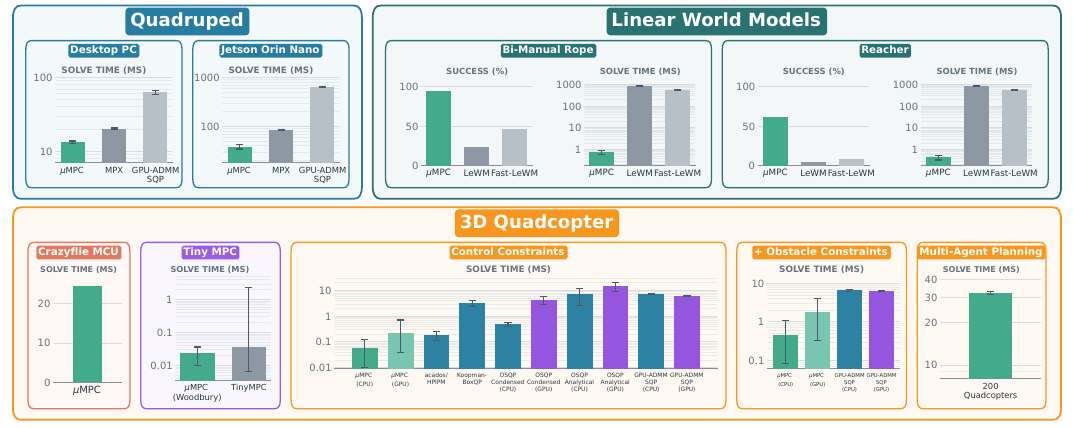}
    \vspace{-1.5em}
    \captionof{figure}{
    $\mu$MPC performance across model, hardware, and problem scale: low-latency
    quadruped and visual-world-model control, onboard microcontroller MPC,
    constrained 3D quadrotor control, and real-time multi-agent planning. $\mu\text{MPC}$ (Woodbury) reports unconstrained Real-Time Iteration (RTI) times for comparisons against Tiny-MPC.
    Solve-time axes are logarithmic where indicated to accommodate the wide
    latency range across methods.
    }
    \label{fig:time_barchart} \vspace{-20pt}
\end{figure*}

\section{Related Work}

\looseness=-1
\noindent\textbf{Koopman MPC and fast optimal control.}
Koopman methods lift nonlinear dynamics into approximately linear coordinates for linear MPC
\cite{korda2018linear,shi2022deepkoopman}, with recent extensions to robotic systems
\cite{li2025continual,jung2026wholebody}.
Existing formulations retain important online computational overhead:
~\cite{korda2018linear} employ a prescribed lifting basis, which may not adequately represent nonlinear costs and constraints in a form amenable to efficient quadratic programming; 
~\cite{wu2024timecertified} condense Koopman MPC but specialize to input constraints; and
~\cite{wu2026boxqp} eliminates high-dimensional observables but keeps physical states and controls as decision variables under box structure.
More broadly, fast MPC exploits sparsity, condensing, cached factorizations, and specialized solvers \cite{stellato2020osqp,verschueren2022acados,nguyen2024tinympc}, while recent GPU methods parallelize dynamics and optimization \cite{amatucci2026primaldual,fang2026gpusls,du2026gato}.
In contrast, $\mu$MPC compiles Koopman dynamics into an action-only KC-QP. For fixed horizon, quadratic weights, and primal regularization, its precomputed primal operator remains reusable across changing goals and constraints, reducing online constrained optimization to matrix--vector products and projections without refactorization.

\looseness=-1
\noindent\textbf{Differentiable control learning.}
Differentiable optimization and MPC support end-to-end learning through constrained control
\cite{amos2017optnet,agrawal2019differentiable,amos2018differentiable,tao2024difftune}.
KCPO~\cite{retchin2023kcpo} combines deep Koopman dynamics with differentiable MPC under hard box control constraints.
We instead differentiate an efficient control-recovery KC-QP that directly supervises the finite-horizon control response, then pair it with DAgger-style on-policy aggregation.
This efficiently targets control relevant model error and controller-induced distribution shift without differentiating through a full expensive nonlinear MPC solve.

\looseness=-1
\noindent\textbf{World models and latent planning.}
World models plan through learned latent dynamics
\cite{hafner2019planet,hansen2024tdmpc2,hafner2025dreamerv3,zhou2025dinowm,maes2026leworldmodel}.
DINO-WM and LeWorldModel optimize action sequences through repeated latent prediction; Fast-LeWM parallelizes prediction but retains CEM search
\cite{gao2026fastlewm}; learned planners instead amortize refinement from imagined rollouts
\cite{sommer2026reinforced}. Historically, these world model dynamics have been represented with high-dimensional nonlinear functions that are expensive to evaluate.  
$\mu$MPC instead learns \textit{linear} world model dynamics, enabling compiliation of the full-horizon response into a compact KC-QP for sub-millisecond planning, avoiding repeated latent rollouts at planning time while permitting explicit linearized state/task constraints and action bounds.

\section{Problem Statement}
 We consider a discrete-time nonlinear system
\begin{equation}
    x_{k+1}=f(x_k,u_k),
    \qquad
    x_k\in\mathbb{R}^{n_x},\;u_k\in\mathbb{R}^{n_u}.
\end{equation}
At each control step, we seek a horizon-$H$ input sequence
$U=[u_0^\top\;\cdots\;u_{H-1}^\top]^\top$ that drives a task-relevant
output $y_k$ toward a goal $y^{\rm goal}$ while satisfying state, task,
and input constraints. Locally linearizing the nonlinear constraints about
the current nominal trajectory gives
\begin{equation}
\small
\hspace{-5pt}\begin{aligned}
\underset{x_{1:H},\,u_{0:H-1}}{\operatorname{minimize}}\quad &
\sum_{k=1}^{H-1}\|y_k-y^{\rm goal}\|_{Q_{y,k}}^2
+\sum_{k=0}^{H-1}\|u_k\|_{R_k}^2
+\|y_H-y^{\rm goal}\|_{Q_f}^2 \\
\operatorname{subject\ to}\quad &
x_{k+1}=f(x_k,u_k),\quad k=0,\ldots,H-1,\\
&
D_{y,k}y_{k+1}+D_{u,k}u_k+\Delta_k\le b_k,
\quad k=0,\ldots,H-1,\\
&
D_fy_H+\Delta_f\le b_f .
\end{aligned}
\label{eq:nonlinear_ocp}
\end{equation}
Here $y_k=S_yx_k$ selects task-relevant coordinates, with $S_y=I$ recovering the full state. We linearize nonlinear state and task constraints about the nominal trajectory, incorporating the constant terms from their first-order Taylor expansions into $b_k$ and $b_f$. Nonnegative margins $\Delta_k$ and $\Delta_f$ then tighten these bounds to account for neglected higher-order terms and prediction errors from the learned dynamics (Sec.~\ref{subsec:KCQP}). Since \eqref{eq:nonlinear_ocp} remains nonlinear through $f$, our goal is to solve \eqref{eq:nonlinear_ocp} by seeking an approximately linear lifted predictive model that moves expensive algebraic operations needed for optimization offline, leaving only fixed linear algebra and projections online.

\section{Method}

\looseness=-1
Our method proceeds in four stages. First, we learn a nonlinear lifting with control affine linear latent dynamics. Second, we compile its finite-horizon response into precomputed maps and eliminate the latent trajectory from the optimization variables. Third, we solve the resulting action-only KC-QP with a primal--dual method whose primal update operator is fixed for a given model, horizon, objective, and primal regularization, and can therefore be precomputed offline and reused as constraints are relinearized online. Finally, we train the representation with differentiable QP for direct control supervision, and DAgger aggregation to improve its finite-horizon control behavior on the distribution induced by its own controller.

\subsection{Learned Dynamics}
\looseness=-1
\noindent In Koopman models \cite{mezic2026koopmansurvey}, the physical state $x_k$ is encoded as
\begin{equation}
    z_k=\operatorname{enc}_{\theta}(x_k),
    \qquad z_k\in\mathbb{R}^{n_z},
\end{equation}
with latent state $z_k$, learned globally linear latent dynamics
\begin{equation}
    z_{k+1}=Az_k+Bu_k,
    \qquad
    A\in\mathbb{R}^{n_z\times n_z},\;
    B\in\mathbb{R}^{n_z\times n_u},
\label{eq:koopman_dynamics}
\end{equation}
and affine decoder
\begin{equation}
    x_k=Cz_k+c,
    \qquad
    C\in\mathbb{R}^{n_x\times n_z},\;c\in\mathbb{R}^{n_x}.
\label{eq:affine_decoder}
\end{equation}
At each MPC step only the measured physical state is encoded, $z_0=\operatorname{enc}_{\theta}(x_0)$; all subsequent predictions follow from the linear latent rollout and decoding into the
physical state space in \eqref{eq:koopman_dynamics}. 
Hereafter, $x_k$ and $y_k$ denote states and outputs predicted by the learned Koopman dynamics, unless otherwise specified. We explicitly distinguish measured or demonstrated states from model predictions when both appear. 

\subsection{Precomputed Maps}
\noindent To accelerate online computation, all horizon operators that depend only on $A$, $B$, $C$, and $H$ are constructed offline.

\looseness=-1
\subsubsection{Condensed latent and physical prediction}
Define
\begin{equation}
    U=\begin{bmatrix}u_0^\top \cdots u_{H-1}^\top\end{bmatrix}^{\!\top}\!\in\mathbb{R}^{Hn_u},
    \quad
    Z=\begin{bmatrix}z_1^\top \cdots z_H^\top\end{bmatrix}^{\!\top}\!\in\mathbb{R}^{Hn_z}.
\end{equation}
Repeatedly applying \eqref{eq:koopman_dynamics} over a prediction horizon of $H$ yields
\begin{equation}
    Z=\Phi z_0+\Gamma U,
\label{eq:condensed_latent}
\end{equation}
where
\begin{equation}
\small
\Phi=
\begin{bmatrix}
A\\A^2\\\vdots\\A^H
\end{bmatrix},
\qquad
\Gamma=
\begin{bmatrix}
B&0&\cdots&0\\
AB&B&\cdots&0\\
\vdots&\vdots&\ddots&\vdots\\
A^{H-1}B&A^{H-2}B&\cdots&B
\end{bmatrix}.
\label{eq:phi_gamma}
\end{equation}
Thus, the latent trajectory is affine in only $(z_0,U)$. Stacking $X=[x_1^\top\;\cdots\;x_H^\top]^\top$ and applying \eqref{eq:affine_decoder} yields
\begin{equation}
    X=a_x(z_0)+G_xU
\label{eq:physical_map}
\end{equation}
with,
\begin{equation}\label{eq:gx}
\begin{aligned}
    a_x(z_0)=(I_H\!\otimes C)\Phi z_0+\mathbf{1}_H\!\otimes c, \quad G_x=(I_H\!\otimes C)\Gamma.
\end{aligned}
\end{equation}
Here, $\otimes$ denotes the Kronecker product, $I_H$ is the $H\times H$ identity matrix, and $\mathbf{1}_H\in\mathbb{R}^H$ is the vector of ones. Note that neither $Z$ nor $X$ must appear as optimization variables since they are defined implicitly through $U$ in \eqref{eq:condensed_latent} and \eqref{eq:physical_map}. 
In implementation, we avoid constructing the full latent prediction matrices $\Phi$ and $\Gamma$. Instead, we directly assemble the physical-state prediction maps from the blocks $CA^k$ and $CA^kB$, for $k=0,\ldots,H$, reducing the need to store intermediate high-dimensional latent trajectories.

\subsubsection{Finite horizon output maps}
Let $S_y\in\mathbb{R}^{n_y\times n_x}$ select the physical-state coordinates relevant to the control objective or constraints, so that $y_k=S_yx_k$. Define 
\begin{equation}
    C_y=S_yC,\qquad c_y=S_yc,\qquad y_k=C_yz_k+c_y.
\end{equation}
For $Y=[y_1^\top\;\cdots\;y_H^\top]^\top$
\begin{equation}
    Y=a_y(z_0)+G_yU
\label{eq:projected_map}
\end{equation}
with,
\begin{equation}
\begin{aligned}
    &a_y(z_0)=F_yz_0+\mathbf{1}_H\!\otimes c_y, \quad F_y=(I_H\!\otimes C_y)\Phi, \\
    &G_y=(I_H\!\otimes C_y)\Gamma.
\end{aligned}
\end{equation}
For each choice of task-relevant outputs $y_k=S_yx_k$, we precompute $F_y$ and $G_y$ offline using the learned dynamics, so that online predictions depend only on the initial latent state $z_0$ and control sequence $U$. 

\subsection{Koopman Compiled Quadratic Program (KC-QP)}
\label{subsec:KCQP}

\subsubsection{Linear surrogate OCP}
We replace the nonlinear rollout in \eqref{eq:nonlinear_ocp} by the learned linear dynamics while retaining the locally linearized constraints:
\begin{equation}
\small
\begin{aligned}
\underset{z_{1:H},u_{0:H-1}}{\operatorname{minimize}}
\quad &
\sum_{k=1}^{H-1}
\|y_k-y^{\mathrm{goal}}\|_{Q_{y,k}}^2
+
\sum_{k=0}^{H-1}
\|u_k\|_{R_k}^2
+
\|y_H-y^{\mathrm{goal}}\|_{Q_f}^2
\\
\text{subject to}\quad
&
z_{k+1}=Az_k+Bu_k,
\quad k=0,\ldots,H-1,
\\
&
D_{y,k}y_{k+1}+D_{u,k}u_k+\Delta_k\leq b_k,
\quad k=0,\ldots,H-1,
\\
&
D_fy_H+\Delta_f\leq b_f.
\label{eq:linear_ocp}
\end{aligned}\hspace{-8pt}
\end{equation}
\looseness=-1
The nonnegative margins $\Delta_k,\Delta_f$ tighten linearized constraints for approximation error, combining linearization and model mismatch terms. Since \eqref{eq:linear_ocp} is a QP, substituting the precomputed maps \eqref{eq:projected_map} eliminates $z_{1:H}$ from the decision variables.

\subsubsection{Condensed objective and constraints}
Let $q=Hn_u$, $Y^{\rm goal}=\mathbf{1}_H\otimes y^{\rm goal}$,
$\bar Q_y=\operatorname{blkdiag}(Q_{y,1},\ldots,Q_{y,H-1},Q_f)$,
and $\bar R=\operatorname{blkdiag}(R_0,\ldots,R_{H-1})$.
Substituting in the finite-horizon prediction map \eqref{eq:projected_map} into the objective of \eqref{eq:linear_ocp} yields a quadratic function of only the control sequence $U$:
\begin{align}
    J(U)
    &=\textstyle\frac{1}{2}
    \|a_y(z_0)+G_yU-Y^{\rm goal}\|_{\bar Q_y}^2
    +\frac{1}{2}\|U\|_{\bar R}^2 \nonumber\\
    &=\textstyle\frac{1}{2}U^\top\mathcal H U
    +g^\top U+c,
\label{eq:condensed_cost}
\end{align}
with
\begin{equation}
\begin{aligned}
    \mathcal H &= G_y^\top \bar Q_y G_y+\bar R, \quad
    g = G_y^\top\bar Q_y
    \bigl(a_y(z_0)-Y^{\rm goal}\bigr),\\
    c &= \frac{1}{2}
    \bigl(a_y(z_0)-Y^{\rm goal}\bigr)^\top
    \bar Q_y
    \bigl(a_y(z_0)-Y^{\rm goal}\bigr).
\end{aligned}
\label{eq:hgc}
\end{equation}
Since $c(z_0)$ is independent of the decision variables $U$, it is dropped. For a fixed model, horizon, and cost weights, changing the goal modifies
only the linear term $g$ (and the discarded constant term), while the
Hessian $\mathcal H$ and its factorization remain unchanged. Stacking stage, terminal, and input inequalities gives $\bar D_yY+\bar D_uU\le \bar b-\bar\Delta$. 
Using \eqref{eq:projected_map},
\begin{equation}
\begin{aligned}
    (\bar D_yG_y+\bar D_u)U
    \le \bar b-\bar\Delta-\bar D_ya_y(z_0),
\end{aligned}
\label{eq:condensed_constraints}
\end{equation}
which we can rewrite as $MU\le r(z_0)$ as shorthand, where $M=\bar D_yG_y+\bar D_u$ and $r(z_0)=\bar b-\bar\Delta-\bar D_ya_y(z_0)$. 
Putting the condensed cost \eqref{eq:condensed_cost} and constraints \eqref{eq:condensed_constraints} together, the complete online QP is therefore 
\begin{equation}
\begin{aligned}
    \underset{U\in\mathbb{R}^{q}}{\operatorname{minimize}}\quad &
    \frac{1}{2}U^\top\mathcal H U+g(z_0)^\top U\\
    \operatorname{subject\ to}\quad &MU\le r(z_0).
\end{aligned}
\label{eq:complete_qp}
\end{equation}
Crucially, only $U$ is optimized; the latent and physical trajectories $Z$ and $X$ have been analytically eliminated.

\subsubsection{Fixed Operator primal--dual solve}
We efficiently solve \eqref{eq:complete_qp} using a tailored primal-dual solver. Primal-dual methods \cite{chambolle2011first} seek a saddle point of the Lagrangian, alternating between descent over the primal decision variables and ascent over the dual variables associated with the constraints. Introducing Lagrange multipliers $\lambda\geq 0$ for the inequality constraints in \eqref{eq:complete_qp}, we define the Lagrangian
\begin{equation}
    \mathcal{L}(U,\lambda)=
    \tfrac{1}{2}U^\top\mathcal{H}U+g(z_0)^\top U
    +\lambda^\top(MU-r(z_0)).
\end{equation}
At iteration $j$, we first perform projected gradient ascent on $\mathcal{L}$ with respect to $\lambda$, with step size $\sigma>0$:
\begin{equation}
    \lambda^{j+1}=\left[\lambda^j+\sigma\big(MU^j-r(z_0)\big)\right]_+.
\label{eq:dual_update}
\end{equation}
This increases the multipliers associated with violated constraints, increasing the penalty for their violation in the subsequent primal update, while the elementwise projection $[\cdot]_+$ maintains nonnegativity of the multipliers. We then minimize $\mathcal{L}(U,\lambda^{j+1})$ with respect to $U$, adding a proximal regularization term with $\rho>0$ to stabilize the update by penalizing deviations from the previous iterate:
\begin{equation}\label{eq:primal_update}
\begin{aligned}
    U^{j+1}=\underset{U}{\operatorname{argmin}}\ 
    &\tfrac{1}{2}U^\top\mathcal H U+g(z_0)^\top U
    +(\lambda^{j+1})^\top MU \\[-5pt]
    &+\tfrac{\rho}{2}\|U-U^j\|_2^2,
\end{aligned}
\end{equation}
where we omit terms independent of $U$. Since \eqref{eq:primal_update} is an unconstrained, strongly convex quadratic minimization, its unique solution satisfies the stationarity condition
\begin{equation}
    (\mathcal H+\rho I_q)U^{j+1}
    =\rho U^j-g(z_0)-M^\top\lambda^{j+1}.
\end{equation}
Defining the Hessian of the proximal primal objective as
\begin{equation}
    K=\mathcal H+\rho I_q,
\label{eq:K}
\end{equation}
we obtain the closed-form primal update
\begin{equation}
    U^{j+1}=K^{-1}\!\left[\rho U^j-g(z_0)-M^\top\lambda^{j+1}\right].
\label{eq:primal_update}
\end{equation}
Crucially, $K$ depends only on the learned dynamics, horizon, objective weights, and $\rho$, and is independent of the constraint matrix $M$ and bounds $r(z_0)$. We therefore compute and store $K^{-1}$ offline, reducing the online primal and dual updates to matrix--vector products and elementwise projections. Although SQP relinearization may change $M$ and $r(z_0)$, these changes enter the updates through matrix--vector products and vector additions, without requiring online matrix inversion or factorization. For each linearized KC-QP, we set $\sigma=0.99\,\rho/\|M\|_F^2$, satisfying the step-size condition $\sigma\|M\|_2^2/\rho<1$. 
To assess convergence, we monitor both primal feasibility and changes in the primal and dual iterates using the residuals
\begin{equation}
\begin{aligned}
    r_{\mathrm{feas}}^{j+1}
    &=\|\big(MU^{j+1}-r(z_0)\big)_+\|_2,\\
    r_{\mathrm{fp}}^{j+1}
    &=\left(\rho^2\|U^{j+1}-U^j\|_2^2
    +\sigma^{-2}\|\lambda^{j+1}-\lambda^j\|_2^2\right)^{1/2}.
\end{aligned}
\end{equation}
The former measures constraint violation, while the latter measures convergence to a fixed point of the primal-dual updates. For each linearized KC-QP, we terminate the solver when both residuals fall below prescribed tolerances or when a fixed real-time iteration budget is exhausted. In the latter case, we return the current iterate, which may be suboptimal or violate constraints.

\subsection{Scalability}
The KC-QP decision dimension is $q=Hn_u$, independent of the lifted dimension $n_z$, compared to $Hn_z+Hn_u$ variables in a non-condensed formulation. The fixed primal operator is precomputed offline in $\mathcal O(q^3)$ time with $\mathcal O(q^2)$ storage. Our implementations precompute the dense inverse $K^{-1}$,
so each primal update is a dense matrix-vector product; on
memory-constrained hardware, the fixed operator can instead be
stored through a factorization.
For $m$ inequalities, each primal-dual iteration requires only $\mathcal O(q^2+mq)$ work. With $N_{\mathrm{PD}}$ inner iterations and $N_{\mathrm{SQP}}$ constraint-linearization iterations, the online optimization complexity is
\begin{equation}
    \mathcal O\!\left(
    N_{\mathrm{SQP}}N_{\mathrm{PD}}(q^2+mq)
    \right),
\end{equation}
excluding constraint linearization and model evaluation. These matrix-vector products and projections are GPU-parallelizable, particularly when solving large batches of KC-QPs. However, CPUs often achieve lower latency for small problems due to GPU launch and synchronization overhead.

The lifted dimension enters online through encoding and affine prediction terms which still scale with $n_z$. Thus, increasing representation capacity does not enlarge the KC-QP decision vector or its primal operator, although encoding and prediction-map
evaluation retain a dependence on $n_z$.

\subsection{Training via Differentiable QP (DiffQP) and DAgger}
\looseness=-1
\noindent
We train on demonstrations $(x_0,U,X)$, with
$U=[u_0^\top\;\cdots\;u_{H-1}^\top]^\top$ and
$X=[x_1^\top\;\cdots\;x_H^\top]^\top$. Rolling out the learned dynamics
under $U$ yields $\hat z_{1:H}$ and $\hat x_{1:H}$,
while demonstrated states are encoded as
$z_{1:H}=\operatorname{enc}_{\theta}(X)$. We initialize $A=I$ and $B$
with Xavier initialization to avoid initially amplifying latent dynamics
or control couplings. We use the multi-step losses
\begin{equation}
\begin{aligned}
    \mathcal L_x&\textstyle=\frac{1}{H}\sum_{k=1}^{H}
    \|\hat x_k-x_k\|_2^2,\quad
    \mathcal L_z=\frac{1}{H}\sum_{k=1}^{H}
    \|\hat z_k-z_k\|_2^2.
\end{aligned}
\label{eq:rollout_losses}
\end{equation}
To supervise finite-horizon control response, we solve
\begin{equation}
    \widehat U
    =
    \underset{V}{\operatorname{argmin}}\;
    \frac{1}{2}
    \left\|a_x(z_0)+G_xV-X\right\|_{\bar Q_{\mathrm{track}}}^2
    +
    \frac{1}{2}\|V\|_{\bar R_{\mathrm{track}}}^2,
\label{eq:diff_qp}
\end{equation}
and optimize
\begin{equation}
    \mathcal L_u=\|\widehat U-U\|_2^2,\qquad
    \mathcal L=\lambda_x\mathcal L_x+
    \lambda_z\mathcal L_z+\lambda_u\mathcal L_u.
\label{eq:training_loss}
\end{equation}
Here, $\bar Q_{\mathrm{track}}\succeq0$ weights deviations from the demonstrated state trajectory, while $\bar R_{\mathrm{track}}\succ0$ regularizes the control sequence. Positive control regularization makes \eqref{eq:diff_qp} strongly convex
with a unique solution. 
By differentiating through the optimal control sequence $\widehat U$, we train the encoder and dynamics parameters $(A,B,C,c)$ so that optimizing the learned model recovers the demonstrated control inputs. 
Crucially, the objective is goal and constraint independent, training the dynamics to recover the demonstrated controls required to reproduce $X$.

\looseness=-1
To mitigate controller-induced distribution shift, we use DAgger-style aggregation. At each epoch, the finite-horizon tracking QP used for control-recovery supervision generates control sequences that are executed in simulation. The resulting trajectories are added to a bounded replay buffer and mixed with original demonstrations, adapting the model to controller-visited states while retaining expert data.

\section{Experiments}
\looseness=-1
Unless otherwise noted, models are trained on an NVIDIA RTX 4090 without state normalization, using loss weights $\lambda_x=1$, $\lambda_z=0.1$, and $\lambda_u=10^{-2}$ in \eqref{eq:training_loss}. Desktop experiments use an i9-14900KF and RTX 4090 with JAX dense linear algebra. unless stated otherwise. Reported times are end-to-end MPC latency. We evaluate three claims: (i) KC-QP reduces constrained MPC latency across desktop, edge, and MCU hardware; (ii) DiffQP and DAgger improve finite-horizon control quality; and (iii) the formulation scales to high-dimensional whole-body and visual models and batched collision-constrained multi-agent planning. For constrained MPC, the solver terminates when feasibility and fixed-point residuals meet prescribed tolerances or the maximum iteration count is reached. 

\subsection{Dubins Car}
\looseness=-1
We first consider a Dubins car \cite[Eq. 13.17]{lavalle2006planning} with state
$x=[p_x,p_y,\sin\theta,\cos\theta]\in\mathbb{R}^4$ and control
$u=[v,\omega]\in\mathbb{R}^2$. We learn a 32-dimensional Koopman model using a three-hidden-layer, width-32 GELU encoder, trained for 250 epochs on 55-step rollouts with batch size 8192 using AdamW. Control-constrained Dubins MPC requires
$0.054\pm0.018/0.066\pm0.019/0.090\pm0.028$ ms on CPU for
$H=15/30/45$, respectively, while GPU latency is
$1.426\pm0.049/1.436\pm0.053/1.438\pm0.058$ ms, dominated by launch overhead for these small action-space QPs.

\subsubsection{Real-Time Multi-Agent Planning}
\looseness=-1
We deploy the learned model on seven differential-drive robots in a $3.2\,\mathrm{m}\times2.0\,\mathrm{m}$ workspace (Fig.~\ref{fig:front_page}F). Each robot solves an $H=28$ KC-QP with the other robots' predicted trajectories as time-varying collision constraints. We perform five iterative best response (IBR) \cite{zhan2026robustly} rounds per control update, with each KC-QP terminating at residual tolerance or after 25 iterations; if the budget is exhausted, IBR proceeds with the current iterate. Using batched NumPy on CPU, the full five-round update takes $18.24 \pm 3.96$ ms, within the $30\,\mathrm{Hz}$ control period. This demonstrates real-time KC-QP solves under changing inter-agent constraints using only CPU-based linear algebra, without GPU acceleration or specialized optimization solvers. 

\subsection{2D Quadcopter}
\looseness=-1
We next consider a planar quadcopter \cite[Eq. 13]{nath2026scalable} with state $x=[p_x,p_z,\theta,v_x,v_z,\omega]$ and inputs given by
collective thrust and torque in a $5\,\mathrm{m}\times5\,\mathrm{m}$
workspace. We learn a 128-dimensional Koopman embedding using a
three-hidden-layer, width-64 GELU encoder. The model
is trained for 250 epochs using 40-step rollouts and a batch size of 8192.

\begin{table}[!htbp]
    \centering
    \caption{Median solve times (ms)}
    \label{tab:quad2d-solve-times}
    \scriptsize
    \setlength{\tabcolsep}{2pt}
    \renewcommand{\arraystretch}{1.12}
    \begin{tabularx}{\columnwidth}{
        >{\raggedright\arraybackslash}X
        c c c}
        \toprule
        \textbf{Method} & \textbf{$H=35$} & \textbf{$H=45$} & \textbf{$H=55$} \\
        \midrule
        \multicolumn{4}{c}{\textit{Control constraints}} \\
        \midrule
        \textbf{$\mu\text{MPC}$ (CPU)}
        & $\mathbf{0.023 \pm 0.027}$
        & $\mathbf{0.033 \pm 0.032}$
        & $\mathbf{0.040 \pm 0.065}$ \\
        $\mu\mathrm{MPC}$ (GPU)
        & $0.129 \pm 0.051$
        & $2.076 \pm 0.845$
        & $2.082 \pm 0.874$ \\
        \midrule
        \multicolumn{4}{c}{\textit{Control + obstacle constraints}} \\
        \midrule
        \textbf{$\mu\text{MPC}$ (CPU)}
        & $\mathbf{0.277 \pm 0.115}$
        & $\mathbf{0.295 \pm 0.075}$
        & $\mathbf{0.322 \pm 0.099}$ \\
        $\mu\mathrm{MPC}$ (GPU)
        & $0.996 \pm 0.487$
        & $0.988 \pm 0.265$
        & $0.949 \pm 0.150$ \\

        \bottomrule
    \end{tabularx}
\end{table}

\looseness=-1
Table~\ref{tab:quad2d-solve-times} shows that $\mu$MPC achieves
$23$--$40\,\mu$s median CPU solve times with control constraints across
the tested horizons. For these small condensed KC-QPs, GPU execution is
slower because launch and synchronization overhead dominate the solve. 
With a centered $1\,\mathrm{m}$-radius obstacle and start--goal pairs on opposing sides, we use an empirical model-error margin. Each of two outer updates relinearizes the nonlinear constraints about the previous control-induced trajectory; the inner primal--dual solve terminates when both residuals fall below $10^{-2}$ or after 150 iterations. Median CPU solve time remains below $0.33$ ms across all horizons while meeting the prescribed residual tolerance.


\subsection{3D Quadcopter}\label{sec:3dquad}

\looseness=-1
We consider a 12-state quadcopter \cite[Eq. 14]{nath2026scalable} with four control inputs in a
$5\,\mathrm{m}\times5\,\mathrm{m}\times5\,\mathrm{m}$ workspace. We use a
128-dimensional Koopman embedding with a three-hidden-layer, width-64 GELU
encoder, trained for 250 epochs with batch size 4096 and
75-step rollouts.

\looseness=-1
We compare against acados/HPIPM \cite{verschueren2022acados},
Koopman-BoxQP \cite{wu2026boxqp}, OSQP \cite{stellato2020osqp} using both
learned Condensed Koopman and analytical dynamics, and GPU-ADMM SQP
\cite{fang2026gpusls}. Unlike trajectory-space SQP methods that update local dynamics online via relinearization, $\mu$MPC reuses a fixed precomputed primal operator across MPC steps and constraint relinearizations. This is enabled by our linear Koopman dynamics and avoids online refactorization.
\begin{table}[!htbp]
    \centering
    \caption{Median solve times (ms)}
    \label{tab:quad3d-solve-times}
    \scriptsize
    \setlength{\tabcolsep}{2pt}
    \renewcommand{\arraystretch}{1.12}
    \begin{tabularx}{\columnwidth}{
        >{\raggedright\arraybackslash}X
        c c c}
        \toprule
        \textbf{Method} & \textbf{$H=50$} & \textbf{$H=75$} & \textbf{$H=100$} \\
        \midrule
        \multicolumn{4}{c}{\textit{Control constraints}} \\
        \midrule
        \textbf{$\mu \text{MPC}$ (CPU)} & $\mathbf{0.058 \pm 0.070}$ & $\mathbf{0.102 \pm 0.092}$ & $\mathbf{0.128 \pm 0.102}$ \\
        $\mu \text{MPC}$ (GPU) & $0.213 \pm 0.496$  & $0.210 \pm 0.161$  & $0.189 \pm 0.149$  \\
        acados/HPIPM  & $0.183 \pm 0.070$ & $0.237 \pm 0.087$ & $0.289 \pm 0.118$ \\
        Koopman-BoxQP & $3.218 \pm 0.830$ & $5.568 \pm 1.416$ & $8.229 \pm 1.706$ \\
        OSQP Condensed (CPU)   & $0.476 \pm 0.078$  & $0.975 \pm 0.069$  & $2.488 \pm 0.243$  \\
        OSQP Condensed (GPU)  & $4.312 \pm 1.557$  & $4.579 \pm 2.745$  & $6.685 \pm 3.046$  \\
        OSQP Analytical (CPU)  & $7.348 \pm 4.676$  & $15.408 \pm 5.025$ & $31.345 \pm 9.291$ \\
        OSQP Analytical (GPU) & $14.510 \pm 5.488$ & $17.634 \pm 6.762$ & $24.532 \pm 9.182$ \\
        GPU-ADMM SQP (CPU)       & $7.239 \pm 0.365$
        & $9.577 \pm 0.392$
        & $13.877 \pm 0.679$ \\
        GPU-ADMM SQP (GPU)      & $6.201 \pm 0.174$
        & $6.351 \pm 0.182$
        & $6.276 \pm 0.103$ \\
        \midrule
        \multicolumn{4}{c}{\textit{Control + obstacle constraints}} \\
        \midrule
        \textbf{$\mu \text{MPC}$ (CPU)}  & $\mathbf{0.454 \pm 0.597}$  & $\mathbf{0.897 \pm 1.047}$  & $\mathbf{1.339 \pm 1.869}$  \\
        $\mu \text{MPC}$ (GPU) & $1.771 \pm 2.233$  & $1.654 \pm 2.258$  & $1.365 \pm 2.462$  \\
        GPU-ADMM SQP (CPU)       & $6.569 \pm 0.335$
        & $9.508 \pm 0.181$
        & $14.051 \pm 0.548$ \\
        GPU-ADMM SQP (GPU)      & $6.174 \pm 0.195$
        & $6.345 \pm 0.158$
        & $6.207 \pm 0.071$ \\
        \bottomrule 
    \end{tabularx}
\end{table}
Table~\ref{tab:quad3d-solve-times} shows $58$--$128\,\mu$s median CPU
latency with control constraints. At $H=100$, $\mu$MPC is $2.3\times$ faster than acados/HPIPM,
$19\times$ faster than OSQP on the identical KC-QP, $64\times$ faster
than Koopman-BoxQP, and over $100\times$ faster than GPU-ADMM SQP on CPU.
The comparison with OSQP using our condensed dynamics isolates the contribution of our solver, which exploits the fixed primal operator to substantially reduce latency beyond condensation alone. 

\looseness=-1
For obstacle avoidance, we introduce a spherical obstacle of radius $1.25\,\mathrm{m}$ and tighten the collision-avoidance constraints using an empirically determined margin to account for model prediction errors, though this can also be calibrated to achieve probabilistic coverage guarantees \cite{srinivasan2026safety}. At each MPC step, we linearize the obstacle constraints twice about the latest predicted trajectory, allowing up to 150 primal-dual iterations per KC-QP solve. Median CPU solve times remain below $1.4$ ms for horizons up to $H=100$, while satisfying the prescribed feasibility tolerance for the linearized constraints. 

\subsubsection{Edge Deployment}
\looseness=-1
We next deploy the same obstacle-aware controller on a Raspberry Pi~5 and
an NVIDIA Jetson Orin Nano to evaluate embedded execution.
\begin{table}[!htbp]
    \centering
    \caption{Median solve times (ms) on edge devices}
    \label{tab:quad3d-edge-solve-times}
    \scriptsize
    \setlength{\tabcolsep}{2pt}
    \renewcommand{\arraystretch}{1.12}
    \begin{tabularx}{\columnwidth}{
        >{\raggedright\arraybackslash}X
        c c c}
        \toprule
        \textbf{Device} & \textbf{$H=50$} & \textbf{$H=75$} & \textbf{$H=100$} \\
        \midrule
        \multicolumn{4}{c}{\textit{Control constraints}} \\
        \midrule
        Raspberry Pi 5 & $0.576 \pm 0.203$ & $1.506 \pm 0.258$ & $3.102 \pm 0.529$ \\
        \textbf{Jetson (CPU)}    & $\mathbf{0.448 \pm 0.297}$ & $\mathbf{0.820 \pm 0.431}$ & $\mathbf{1.314 \pm 0.756}$ \\
        Jetson (GPU)    & $1.760 \pm 1.032$ & $1.823 \pm 0.800$ & $1.860 \pm 0.770$ \\
        \midrule
        \multicolumn{4}{c}{\textit{Control + obstacle constraints}} \\
        \midrule
        Raspberry Pi 5 & $7.255 \pm 2.021$ & $25.488 \pm 3.049$ & $42.590 \pm 4.679$ \\
        Jetson (CPU)    & $\mathbf{2.369 \pm 2.430}$ & $\mathbf{4.700 \pm 4.418}$  & $7.109 \pm 8.987$  \\
        Jetson (GPU)    & $5.213 \pm 4.768$ & $4.967 \pm 4.033$  & $\mathbf{4.470 \pm 4.530}$  \\
        \bottomrule
    \end{tabularx}
\end{table}
Table~\ref{tab:quad3d-edge-solve-times} shows that control-constrained MPC
remains below $3.2$ ms on every edge configuration through $H=100$.
With obstacle constraints, the Jetson CPU remains below $7.2$ ms, while
the GPU becomes advantageous at longer horizons as the larger constrained
problem better amortizes device overhead.

\subsubsection{Differentiable-QP Ablation}
\looseness=-1
We isolate the contributions of DAgger and differentiable-QP supervision, with numerical results in Table \ref{tab:diffqp-justification}.
\begin{table}[!htbp]
    \centering
    \caption{Differentiable QP and DAgger ($\dagger$) ablation. Horizon
entries report goal-reaching success (\%) / MPC steps to goal.}
    \label{tab:diffqp-justification}
    \scriptsize
    \setlength{\tabcolsep}{2.2pt}
    \renewcommand{\arraystretch}{1.15}
    \begin{tabularx}{\columnwidth}{
        >{\raggedright\arraybackslash}X
        c c c}
        \toprule
        \textbf{Method}
        & $H=75$
        & $H=100$
        & $H=125$ \\
        \midrule
        Base
        & 3.71 / 155.5
        & 6.84 / 160.5
        & 6.84 / 215.0 \\

        Base $+\dagger$
        & 90.82 / 343.25
        & 80.27 / 290.75
        & 47.07 / 277.25 \\

        Base $+$ diffQP
        & \textbf{97.66} / 328.5
        & 92.19 / 516.6
        & 85.55 / 508.7 \\

        Base $\mathbf{+\dagger+}$ diffQP
        & 97.46 \textbf{/ 299.0}
        & \textbf{96.88 / 266.5}
        & \textbf{95.70 / 285.5} \\
        \midrule
        & \(\kappa(\bar Q^{1/2}G_x)\downarrow\)
        & \(\sigma_{\min}(\bar Q^{1/2}G_x)\uparrow\)
        & Control RMSE \(\downarrow\) \\
        \midrule
        Base
        & $6666$
        & $2.73e{-4}$
        & $0.814$ \\
        
        Base $+\dagger$
        & $8612$
        & $2.16e{-4}$
        & $0.813$ \\

        Base $+$ diffQP
        & $3551$
        & $4.82e{-4}$
        & $\mathbf{0.747}$ \\
        
        Base $\mathbf{+\dagger+}$ diffQP
        & $\mathbf{1779}$
        & $\mathbf{1.03{\times}10^{-3}}$
        & $0.787$ \\
        \bottomrule
    \end{tabularx}
\end{table}
\looseness=-1
To assess how differentiable QP supervision affects the learned control-to-state map $G_x$ in \eqref{eq:gx}, we report the condition number $\kappa_2(\bar Q_{\mathrm{track}}^{1/2}G_x)$ and smallest singular value $\sigma_{\min}(\bar Q_{\mathrm{track}}^{1/2}G_x)$, where $\bar Q_{\mathrm{track}}$ is the state-tracking weight matrix in \eqref{eq:diff_qp}.
A smaller $\kappa_2$ indicates more uniform sensitivity of the predicted state trajectory to perturbations along different control directions, while a larger $\sigma_{\min}$ indicates fewer near-degenerate control directions. Both improve the numerical conditioning of recovering control inputs from desired state trajectories. 
DiffQP raises success from $3.7\%$ to $97.7\%$
at $H=75$ and retains $85.6\%$ at $H=125$, while improving conditioning
and control-recovery RMSE. Adding DAgger maintains over $95\%$ success
through $H=125$ and reaches the goal in fewer MPC steps at long horizons.
Although the combined model has slightly higher control RMSE than DiffQP
alone, it has substantially better closed-loop performance, showing that DAgger complements DiffQP by adapting
the model to controller-induced states. At batch size 4096, DiffQP incurs only modest additional training overhead, increasing per-epoch time from 21.97 to 26.82 s and peak VRAM from 2082 to 2224 MiB.

\subsubsection{Comparison with TinyMPC}
\looseness=-1
For the unconstrained KC-QP, the optimal control sequence is given by $U^\star=-\mathcal H^{-1}g(z_0)$, where $\mathcal H=\bar R+\tilde G^\top\tilde G$ and $\tilde G=\bar Q_y^{1/2}G_y$ \eqref{eq:hgc}. When $\tilde G$ has fewer rows than columns, the Woodbury matrix identity \cite{henderson1981deriving} replaces the inversion of the $Hn_u\times Hn_u$ Hessian with that of the smaller matrix $I+\tilde G\bar R^{-1}\tilde G^\top$. Since these matrices are fixed and can be precomputed offline, the optimal control sequence can be recovered exactly using only matrix--vector products online. 
\begin{table}[!htbp]
    \centering
    \caption{Mean / median $\pm$ std.\ RTI latency (ms) versus TinyMPC}
    \label{tab:quad3d-solve-time}
    \scriptsize
    \setlength{\tabcolsep}{1.5pt}
    \renewcommand{\arraystretch}{1.12}
    \begin{tabularx}{\columnwidth}{
        >{\raggedright\arraybackslash}X
        c c c}
        \toprule
        \textbf{Method}
        & \textbf{$H=50$}
        & \textbf{$H=60$}
        & \textbf{$H=70$} \\
        \midrule
        \mbox{\textbf{$\mu\text{MPC}$ (Woodbury)}}
        & $\mathbf{0.029 / 0.023 \pm 0.013}$
        & $\mathbf{0.035 / 0.029 \pm 0.017}$
        & $\mathbf{0.045 / 0.037 \pm 0.019}$ \\
        TinyMPC
        & $0.853 / 0.036 \pm 2.280$
        & $0.722 / 0.038 \pm 2.188$
        & $0.785 / 0.042 \pm 2.380$ \\
        \bottomrule
    \end{tabularx}
\end{table}
The Woodbury form requires $23$--$37\,\mu$s median latency, comparable
to TinyMPC's median runtime but with substantially lower mean latency and
variance. We note that the two methods solve different MPC problems: $\mu$MPC solves an unconstrained QP using learned Koopman dynamics that can capture the nonlinear dynamics, whereas TinyMPC enforces actuator bounds and uses dynamics linearized analytically about hover, with a different terminal cost. Thus, Table~V compares the computational cost of executing each MPC controller, rather than the efficiency of the two solvers on the same optimization problem. 

\subsubsection{Crazyflie Deployment}
\looseness=-1
We deploy $\mu$MPC on a Crazyflie Cortex--M4F for the point-to-point task in Fig.~\ref{fig:front_page}B using a 128-dimensional Koopman model with $H=75$ ($q=300$, $1.875\,\mathrm{s}$ horizon). This deployment uses an unconstrained KC-QP, so the optimal action sequence is obtained directly from the fixed quadratic system without primal--dual iterations. For memory efficiency, the MCU stores precompiled \texttt{float32} quantities and a packed Cholesky factor. The controller requires $601.7\,\mathrm{kB}$ of immutable storage and $8.7\,\mathrm{kB}$ of runtime workspace; total firmware size is $948.4\,\mathrm{kB}$.

After hover-thrust calibration, MPC commands normalized thrust and body rates. The MCU averages $20.08\,\mathrm{ms}$ over 100 benchmark cases with no $25\,\mathrm{ms}$ deadline violations and $24.57\,\mathrm{ms}$ during flight. The full radio-mediated experimental loop runs at approximately $15\,\mathrm{Hz}$ due to communication, state acquisition, Vicon forwarding, and telemetry.

\subsubsection{Scalable Multi-Agent Quadcopter Planning}
\looseness=-1
We employ Jacobi IBR for multi-agent planning in a $25\,\mathrm{m}$ cube; that is, at each IBR iteration, each quadrotor optimizes its trajectory while treating the other agents' trajectories from the previous iteration as fixed, time-varying collision constraints. This enables all agents' KC-QPs to be solved simultaneously in a single GPU batch. We reuse the original Koopman model from Sec. \ref{sec:3dquad} without retraining by planning in moving $5\mathrm{m}$ local frames centered on each robot. 
We use up to $2$ IBR sweeps, $1$ SQP iteration, and $150$ PD iterations per KC-QP, terminating early upon convergence and accepting the current iterate when the budget is exhausted. Goal \% denotes agents
reaching their assigned goal; Collision \% denotes agents experiencing at
least one collision.

\begin{table}[!htbp]
    \centering
    \caption{GPU-Batched, Multi-Quadrotor MPC at Scale}
    \label{tab:quad3d-robot-scaling}
    \scriptsize
    \setlength{\tabcolsep}{3pt}
    \renewcommand{\arraystretch}{1.12}
    \begin{tabularx}{\columnwidth}{
        >{\raggedright\arraybackslash}X
        c c c c}
        \toprule
        \textbf{Metric}
        & 50 Quadrotors
        & 100 Quadrotors
        & 150 Quadrotors
        & 200 Quadrotors \\
        \midrule
        \mbox{Solve (ms)}
        & $12.089 \pm 0.017$
        & $17.347 \pm 0.156$
        & $23.020 \pm 0.754$
        & $32.117 \pm 0.913$ \\
        \mbox{Goal \%}
        & $100.0 \pm 0.0\%$
        & $99.760 \pm 0.427\%$
        & $99.173 \pm 0.633\%$
        & $98.420 \pm 1.137\%$ \\
        \mbox{Collision \%}
        & $0.0 \pm 0.0\%$
        & $1.240 \pm 1.305\%$
        & $1.867 \pm 1.373\%$
        & $2.580 \pm 1.278\%$ \\
        \bottomrule
    \end{tabularx}
\end{table}
Scaling from $50$ to $200$ quadrotors increases batched MPC/IBR latency
from $12.1$ to $32.1$ ms ($2.66\times$ for $4\times$ more agents), while
maintaining over $30\,\mathrm{Hz}$ operation and above $98\%$ goal
reaching. Collision incidence rises to $2.58\%$ at $D=200$. At large
scales, the fixed iteration budget trades full best-response convergence
for real-time updates, so residual collision-constraint violations may
remain. KC-QP is well suited to IBR because inter-agent constraints change
each round while the primal operator remains fixed.

\subsection{Quadruped}
\subsubsection{Control Constrained Planning}
\looseness=-1
We evaluate whole-body quadruped control using a 4096-dimensional Koopman model of the 61-dimensional robot state, comprising floating-base pose and velocity, joint states, foot positions, and ground-reaction forces, with 12 joint-torque inputs. The three-hidden-layer, width-512 GELU encoder is trained for 100 epochs with batch size 2500 and 55-step rollouts. To avoid out-of-distribution analytical gait references, we distill a PPO locomotion policy into a learned reference generator supplying joint states, foot heights, and vertical ground-reaction forces for KC-QP tracking. Fixed base height and commanded linear and angular velocities are also tracked. We compare against MPX's primal--dual iLQR~\cite{amatucci2026primaldual} and GPU-ADMM SQP~\cite{fang2026gpusls} using identical reference generation and torque limits. Reported latencies include reference generation for all methods and state encoding for $\mu$MPC.
\begin{table}[!htbp]
    \centering
    \caption{Median solve times (ms)}
    \label{tab:go2-solve-time}
    \scriptsize
    \setlength{\tabcolsep}{1.5pt}
    \renewcommand{\arraystretch}{1.12}
    \begin{tabularx}{\columnwidth}{
        >{\raggedright\arraybackslash}X
        c c c}
        \toprule
        \textbf{Method}
        & \textbf{$H=25$}
        & \textbf{$H=65$}
        & \textbf{$H=100$} \\
        \midrule
        \multicolumn{4}{c}{\textit{Control constraints}} \\
        \midrule
        \textbf{$\mu \text{MPC}$}
        & $\mathbf{4.472 \pm 0.253}$
        & $\mathbf{8.760 \pm 0.356}$
        & $\mathbf{13.295 \pm 0.456}$ \\
        MPX
        & $6.793 \pm 0.331$
        & $13.464 \pm 0.483$
        & $20.062 \pm 0.555$ \\
        GPU-ADMM SQP
        & $23.743 \pm 1.459$
        & $47.111 \pm 2.560$
        & $62.951 \pm 3.821$ \\
        \midrule
        \multicolumn{4}{c}{\textit{Deployment on Jetson with Control Constraints}} \\
        \midrule
        \textbf{$\mu \text{MPC}$}
        & $\mathbf{24.467 \pm 1.887}$
        & $\mathbf{32.185 \pm 3.231}$
        & $\mathbf{38.110 \pm 3.955}$ \\
        MPX
        & $28.437 \pm 1.550$
        & $70.881 \pm 3.269$
        & $84.197 \pm 2.992$ \\
        GPU-ADMM SQP
        & $173.023 \pm 6.405$
        & $409.777 \pm 6.475$
        & $637.463 \pm 8.696$ \\
        \bottomrule
    \end{tabularx} \vspace{-2pt}
\end{table}
\begin{figure}[!htbp]
    \centering
    \includegraphics[width=\columnwidth]{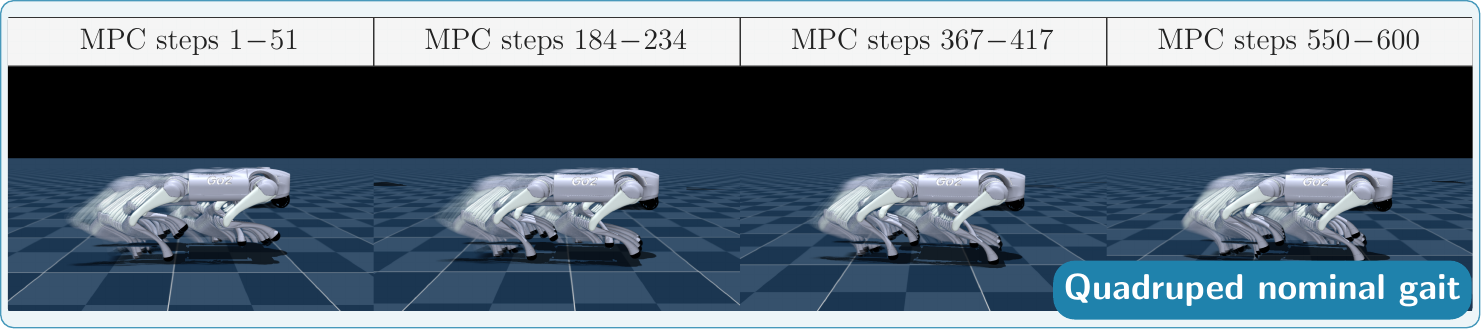}
    \caption{Nominal Go2 locomotion under $\mu$MPC over a closed-loop rollout.}
    \label{fig:go2_nom} \vspace{-2.5pt}
\end{figure}

\looseness=-1
Table~\ref{tab:go2-solve-time} shows that at $H=100$, $\mu$MPC is $1.5\times$ faster than MPX and $4.7\times$ faster than GPU-ADMM SQP; the shared reference generator contributes $2.71/5.72/8.53$ ms at $H=25/65/100$. On the Nvidia Jetson Orin Nano, $\mu$MPC runs in $24$--$38$ ms, versus $28$--$84$ ms for MPX and $173$--$637$ ms for GPU-ADMM SQP. All control-constrained solves terminate within the 150-iteration
primal--dual budget at the standard $10^{-2}$ residual tolerance.

\subsubsection{Whole-Body Obstacle-Avoidance Filtering}
\looseness=-1
We filter the same PPO policy, operating within a learned simulator, for state/torque reference and KC-QP obstacle constraints. This harder
constrained problem uses up to 750 primal--dual iterations and a relaxed
$5{\times}10^{-2}$ residual tolerance to preserve the real-time rate.
Across our evaluated rollouts, the filtered controller avoids the obstacle
crossed by the nominal policy (Fig.~\ref{fig:front_page}A). At $H=100$,
end-to-end latency is $28.18$ ms: $7.74$ ms reference generation,
$0.22$ ms encoding, and $20.13$ ms KC-QP optimization. Constraint tightenings added for robustness to model mismatch are empirically tuned rather than probabilistically calibrated;
formal robust margins remain future work and can be derived following \cite{srinivasan2026safety} and \cite{li2026robustness}.

\looseness=-1
\subsection{Linear World Models for Visual Planning}

We evaluate all planners on pixel-based bimanual rope manipulation and Reacher \cite[Fig. 2c-d]{nath2026pixels} using Koopman-linearized latent world models for planning and control. We evaluate the efficacy of our planner using random workspace start--goal pairs, rather than LeWM \cite{maes2026leworldmodel} and Fast-LeWM's \cite{gao2026fastlewm} 25-step same-trajectory goal sampling. Images are encoded by a custom ViT into a 12-dimensional embedding, augmented with first- and second-order differences to form
$s_t=[e_t,\Delta e_t,\Delta^2 e_t]\in\mathbb{R}^{36}$, and lifted to a 2048-dimensional Koopman model with linear dynamics. For rope manipulation, $u_t=[w_t,h_t,r_t]$ specifies gripper separation, height, and fore--aft reach and is executed through inverse kinematics. We compile the KC-QP offline, so the 2048-dimensional latent trajectory never appears in the online decision variables; planning optimizes only the action sequence with explicit action bounds enforced as linear inequalities.
\begin{table}[!htbp]
    \centering
    \caption{End-to-end planning performance and solve times (ms)}
    \label{tab:linear-wm-planning}
    \scriptsize
    \setlength{\tabcolsep}{2pt}
    \renewcommand{\arraystretch}{1.12}
    \begin{tabularx}{\columnwidth}{
        >{\raggedright\arraybackslash}X
        c c c}
        \toprule
        \textbf{Metric} & \textbf{$\mu\text{MPC}$} & LeWM & Fast-LeWM \\
        \midrule
        \multicolumn{4}{c}{\textit{Bi-Manual Rope}} \\
        \midrule
        Solve time (ms) & $\mathbf{0.725 \pm 0.164}$ & $820.541 \pm 42.092$ & $559.063 \pm 41.601$ \\
        Success rate    & $\mathbf{94\%}$             & $24\%$                & $46\%$ \\
        \midrule
        \multicolumn{4}{c}{\textit{Reacher}} \\
        \midrule
        Solve time (ms) & $\mathbf{0.427 \pm 0.098}$ & $814.2 \pm 30.9$ & $558.8 \pm 29.2$ \\
        Success rate    & $\mathbf{62\%}$             & $4\%$             & $8\%$ \\
        \bottomrule
    \end{tabularx}\vspace{-5pt}
\end{table}

Table~\ref{tab:linear-wm-planning} shows that $\mu$MPC plans in
$0.725$ ms on bimanual rope manipulation (Fig. \ref{fig:front_page}d) and $0.427$ ms on Reacher, while achieving
$94\%$ and $62\%$ success, compared with $24\%/4\%$ for LeWM and
$46\%/8\%$ for Fast-LeWM. All evaluated KC-QPs reach the prescribed
primal--dual residual tolerances within the real-time iteration budget and respect the imposed action bounds. On the Jetson GPU, the same planners run in approximately $33.060 \pm 1.107$ ms on both tasks, sustaining about $30$ Hz even on edge hardware. These results demonstrate that scaling learned visual representations to 2048 dimensions preserves fast online trajectory optimization via Koopman linearization without requiring repeated expensive nonlinear latent rollouts or sampling-based trajectory search.

\subsubsection{Hardware Cube Manipulation}
\begin{figure}[!htbp]
\vspace{-5.0pt}
    \centering
    \includegraphics[width=\columnwidth]{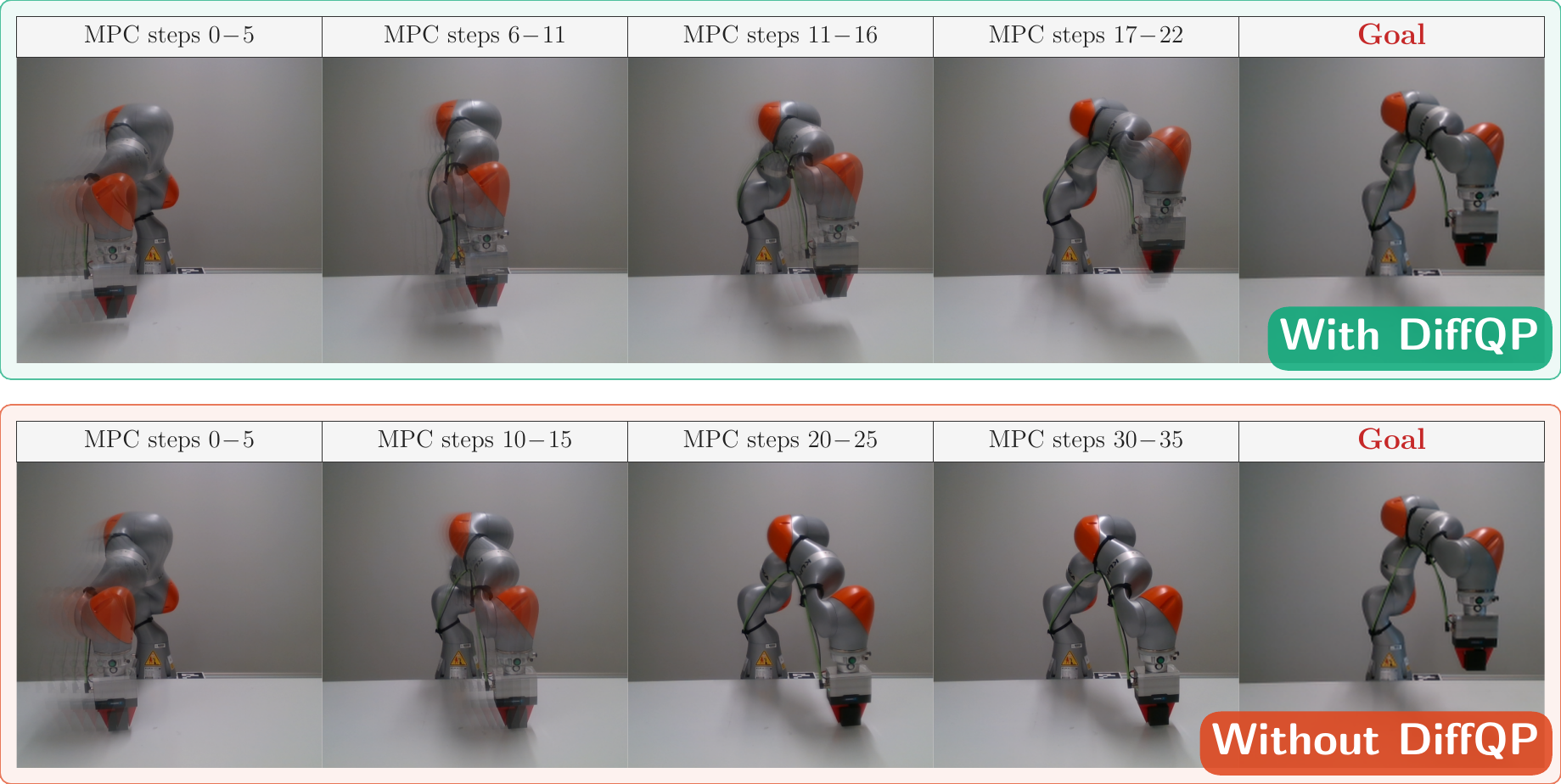}
    \caption{Real-world cube manipulation with and without diffQP
    supervision.}
    \label{fig:cube_real_diffqp_comparison} \vspace{-0pt}
\end{figure}
\looseness=-1
We validate the same formulation on real cube manipulation with a Kuka iiwa arm, where
$\mu$MPC commands Cartesian end-effector targets
$u_t=[x_t,y_t,z_t]$ through inverse kinematics to match a goal image taking $0.483 \pm 0.130$ ms per solve. As shown in
Fig.~\ref{fig:cube_real_diffqp_comparison}, differentiable-QP supervision
recovers the vertical control authority required to reach the elevated target, whereas prediction-only training remains below the goal height despite approaching the correct planar configuration.

\section{CONCLUSION}
\looseness=-1
We presented $\mu$MPC, a framework that combines scalable Koopman dynamics learning with fast constrained MPC. By compiling finite-horizon predictions into action-only KC-QPs with a reusable primal operator, $\mu$MPC reduces online primal-dual optimization to matrix-vector products and projections, invariant under changing goals and constraints. Lightweight differentiable control-recovery QPs enable fast, batched supervision and scalable DAgger-style data aggregation, directly training dynamics for finite-horizon control. Across experiments, $\mu$MPC achieves millisecond-scale whole-body control with 4096-dimensional Koopman models, submillisecond planning with 2048-dimensional linear world models, long horizon MPC onboard Crazyflie MCU, and real-time planning for 200 quadrotors.

\looseness=-1
Future work should focus on principled model selection and improved constraint handling. Our reported encoder capacities, lifted dimensions, and training horizons were selected through empirical tuning and retraining, motivating scaling laws relating these choices to prediction and closed-loop performance. In very demanding cases, fixed-budget solves may retain residual constraint violations, incorporating constraints into training could improve constraint response with fewer primal-dual iterations. Current model-error margins also remain empirical rather than formally calibrated.







\bibliographystyle{IEEEtran}
\bibliography{references}


\end{document}